\documentclass[11pt]{article}
\usepackage[T1]{fontenc}
\usepackage[utf8]{inputenc}
\usepackage{lmodern,microtype,amsmath,booktabs,array,longtable,tabularx}
\usepackage[margin=1in]{geometry}
\usepackage{tikz}
\usetikzlibrary{arrows.meta,positioning}
\usepackage{textcomp,listings,xurl}
\usepackage[hidelinks]{hyperref}
\newcommand{\NA}{---}
\title{Decision Readouts for Text-Mediated Video Anomaly Detection:\\An Exploratory Evaluation of Jev and Qwen}
\author{\small Xukui Qin, Youting Wang, Xinjie He\\\small Ziyang Luo, Runxiong Wu, Yan-Syuan Chen, Zhongyao Chu\\[4pt]\small Independent researchers}
\date{27 September 2026}
\begin{document}
\maketitle
\begin{center}\small Exploratory preprint, version 1\end{center}
\begin{abstract}
How much does the decision readout matter when video-derived textual evidence is held fixed? We evaluate Jev typed decisions and three Qwen readouts on a sparse development sample of 40 videos and 400 target anchors from UCF-Crime and XD-Violence, each presented as a summary and ordered captions. Each dataset contributes 20 source groups and 200 anchors, including only 10 and 37 positives, respectively. The original five-backend pilot requested 4,000 predictions; Jev Choice returned 776 valid responses out of 800 under the study's strict numerical policy, blocking its full-coverage quality comparison. On XD captions, Jev Noul achieved 75.99\% average precision versus 48.47\% for Qwen generated probability and 57.81\% for the stronger local ordinal-likelihood expectation. The latter paired difference was 18.18 percentage points (95\% source-group bootstrap interval 5.53--31.50). UCF did not show a corresponding advantage: caption ROC-AUC was 52.26\% for Noul and 65.95\% for ordinal likelihood. Both probability readouts had higher, hence worse, UCF Brier scores than the evaluation-prevalence reference of 0.0475. We additionally audit historical LAVAD scores at exactly matched anchors and distinguish response structure from numerical consistency. A binary-likelihood control is missing. These exploratory offline results characterize ranking, probability quality and interface failures; they establish neither a causal typed-interface benefit nor general superiority, calibration or end-to-end acceleration.
\end{abstract}

\section{Introduction}
Converting a video to text separates evidence construction from the decision made using that evidence. Once descriptions are fixed, the decision can be a generated rating, a probability number, a likelihood over legal answers, or a response from a typed API. These alternatives differ in output semantics, resolution, computation and failure modes. The scientific question is whether their practical differences remain useful when the information supplied to them is controlled.

We study this question using cached visual descriptions from LAVAD \cite{r1,r2}. We compare short generated ratings, full-answer ordinal likelihoods and generated event probabilities from one fixed Qwen model with Jev Choice and Noul. We evaluate both ranking and agreement with temporal benchmark labels, without treating an ordinal rating as a binary probability. Historical LAVAD scores provide context at exactly the same queried anchors, while a separate full-cache replay checks the evaluation path.

The contribution is an exploratory empirical audit, not a new visual model or likelihood estimator. It consists of a paired text-view pilot, probability references that reveal weak performance under class imbalance, and a response-validity analysis that preserves failed requests. The 40 videos, not the 4,000 predictions, describe the primary sample size. Cross-model comparisons also change training and service implementation; they cannot identify a causal effect of typing an API. Within-Qwen comparisons constrain more factors, but changing a probability-number prompt to a YES/NO answer necessarily changes its output instruction.

\section{Related work}
LAVAD combines visual captioning, caption cleaning, temporal summaries, anomaly scoring and retrieval-based refinement without task-specific training \cite{r1,r2}. UCF-Crime and XD-Violence provide distinct temporal evaluation conventions \cite{r3,r4}; we use their author-cache metadata and labels, not a new annotation study. We do not use audio.

Likelihood readout is established related work. Song and Lee examine rank compression in decoded VLM answers and probability-weighted readouts \cite{r9}. Probe-VAD uses ordinal likelihood probing on visual clips, including ordered binary thresholds \cite{r10}. Our contribution is narrower: fixed cached text, two decision providers, binary-probability diagnostics and explicit numerical rejection accounting. Neither those papers' methods nor their benchmark numbers were reproduced here. Flashback uses memory-driven retrieval to pursue real-time anomaly detection \cite{r11}; our noncausal cached inputs do not support that claim. At the representation level, MDSF combines multi-scale teacher--student discrepancy and saliency fusion in a masked-autoencoder framework \cite{r12}. Its visual learning setting and datasets differ from our decision-stage study, precluding a direct numerical comparison.

TypeSafe's vendor documentation defines Noul as a yes/no probability and Choice as an option with a probability distribution \cite{r5}. Jev was announced on 15 September 2026 \cite{r8}. An output contract is not independent evidence of calibration or better anomaly detection. Following the distinction between stated confidence and empirical correctness \cite{r6}, we evaluate probabilities against labels rather than infer their reliability from their type. The local model is the officially released Qwen3-4B-Instruct-2507 \cite{r7}.

\section{Experimental design}
\subsection{Scope, sampling and temporal targets}
The completed development pilot selected ten normal and ten anomalous videos per dataset using video-level metadata, not target-anchor labels. Within each bucket, videos were ordered by SHA-256 of a fixed seed, dataset name and source identifier; normal videos were selected first, skipping source groups already used. Ten approximately evenly spaced stride-16 anchors per selected video gave 200 anchors per dataset. Appendix~\ref{app:sampling} specifies the exact ordering, grouping, anchor rule and frozen manifest hash. There are 20 source groups per dataset: one video per UCF group; XD groups use the original source-title prefix. The sample contains 10 UCF positive anchors in six groups and 37 XD positive anchors in ten groups. These groups are not verified independent abnormal events. The same 400 anchors appear in both text views; views do not double sample size.

The pilot is a disclosed, nonstandard sparse development sample. Its selection was approved before new scoring, but the present analyses are exploratory. A separately frozen remaining-group study is outside this version's scope. Its partial quality results were not read or included. The full author-cache replay, the pilot and that study remain distinct scopes.

Each request identifies an exact target frame and the available context. LAVAD's nominal ten-second window is centered on the target and clipped at video boundaries, with ten sampled caption positions. The cache assumes 30 FPS for UCF and 24 FPS for XD; these metadata were not independently measured by decoding the source video. Cleaned captions can retrieve descriptions from elsewhere in the video, so their attached frame indices locate query positions and do not prove that every description originated there. Future context and nonlocal retrieval limit every result here to offline analysis.

\subsection{Fixed evidence and changing decisions}
\begin{figure}[htbp]\centering
\begin{tikzpicture}[box/.style={draw,rounded corners,align=center,font=\small,text width=4.7cm,minimum height=1cm},>=Latex]
\node[box] (cache) at (0,0) {Frozen author cache\\visual captions and cleaning};
\node[box] (views) at (0,-2.1) {Frozen paired evidence\\summary / ordered captions\\same target and context};
\node[box] (q) at (-2.75,-4.7) {Qwen: same weights/runtime\\generated rating/probability\\ordinal full-answer likelihood};
\node[box] (j) at (2.75,-4.7) {Jev: different model/service\\Choice distribution / Noul\\typed response validation};
\node[box,text width=7.7cm] (eval) at (0,-7.2) {Closed outputs + isolated benchmark labels\\ranking, probability quality, coverage and paired group intervals};
\draw[->] (cache)--(views);\draw[->] (views)--(q);\draw[->] (views)--(j);\draw[->] (q)--(eval);\draw[->] (j)--(eval);
\end{tikzpicture}
\caption{Controlled and changing components. Labels, source names and historical scores never enter new-scorer requests. The separate historical LAVAD raw reference is joined only for evaluation.}
\label{fig:design}
\end{figure}
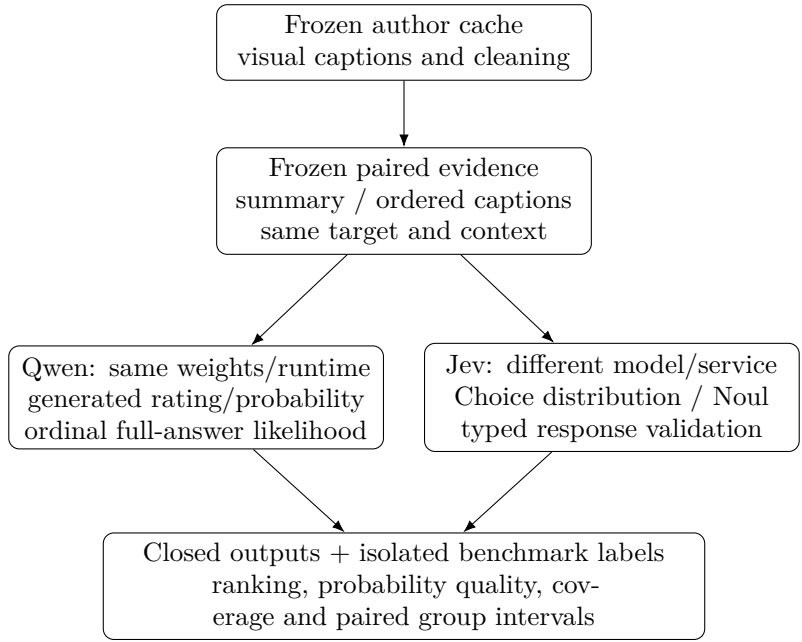

Figure~\ref{fig:design} shows the information boundary. The summary and caption conditions use different amounts and organizations of cached information; this is not a pure formatting intervention. The caption-input path uses the original visual-caption index and cleaning operation, without summary text, summary embeddings or score refinement in the inspected dependency graph. We nevertheless call it \emph{caption-input}, rather than claim independence from all upstream processing. The cleaning code and evidence provenance were inspected; no upstream model was rerun.

The serialization allowlist includes target frame, FPS, context bounds, view and its text. It excludes filenames, class-bearing source names, labels, historical scores and local record IDs. Evaluation labels reside in a separate sidecar. A schematic, explicitly nonempirical example and exact prompt construction appear in Appendix~\ref{app:prompts}.

The unchanged broad event definition is ``Visible physical violence, dangerous incidents, or other clearly abnormal activity.'' This definition does not exactly coincide with each benchmark's annotation policy. Probability scores consequently measure agreement with benchmark labels under this prompt, not verified calibration for an independently validated real-world event definition.

\subsection{Backends and probability semantics}
The original pilot used five backends. Qwen short generation returned one of eleven ratings $A=\{0.0,0.1,\ldots,1.0\}$. Qwen ordinal likelihood scored each complete bracketed answer, normalized over $A$, and produced its expectation and argmax. Qwen generated probability returned a number in $[0,1]$ for the defined binary event. Jev Choice returned the eleven-option distribution, from which the same expectation and argmax were computed. Jev Noul returned an event probability. Ordinal expectations and vendor confidence were excluded from Brier and NLL.

For a legal answer $a$, likelihood scoring used
\begin{equation}\ell(a\mid x)=\sum_{t=1}^{|a|}\log P(a_t\mid x,a_{<t}),\qquad
w(a)=\frac{\exp\ell(a\mid x)}{\sum_{b\in A}\exp\ell(b\mid x)}.\label{eq:likelihood}\end{equation}
All candidate tokens and the EOS terminator were included, without length normalization. Prompt/answer token boundaries were checked after the chat template. A shared prefill was copied into isolated mutable candidate caches and tested against independent full-sequence forwards. Such agreement checks implementation, not prediction correctness.

Qwen used the fixed revision in Appendix~\ref{app:runtime}, NF4 with FP16 computation, SDPA, batch size one and a 2,048-token limit. Generation was greedy with at most 16 new tokens and no explanation. Jev was pinned to \texttt{jev-1.13.0}. These are efficient output baselines, not deliberately verbose chains of thought.

A post-pilot control for full-sequence \texttt{["YES"]}/\texttt{["NO"]} likelihood was implemented and frozen before any new quality evaluation. Its probability is $\exp(\ell_{yes}-\operatorname{logsumexp}(\ell_{yes},\ell_{no}))$. The question and output contract change only as needed to ask for those answers; the exact difference is retained in Appendix~\ref{app:prompts}. This control was not executed in this version. Execution records are not retroactive preregistration. Neither candidate normalization nor Noul's scalar output establishes calibration.

\subsection{Metrics, references and uncertainty}
UCF's primary sampled-anchor metric is ROC-AUC; XD's is standard AP. Trapezoidal PR-AUC is separate, especially because tied ordinal scores can change the relationship between these areas. No unqueried frame is imputed for pilot evaluation. Brier is $n^{-1}\sum_i(p_i-y_i)^2$; NLL uses natural logarithms with each log argument floored at $10^{-15}$. Ten-bin equal-width ECE is descriptive and reported in Table~\ref{tab:probability}, not treated as proof of calibration.

We added constant $p=0$ and $p=\pi$, where $\pi$ is the evaluation sample's positive fraction. The latter is an \emph{evaluation-prevalence descriptive reference}: it uses evaluation labels after scoring, is not a deployable learned prior, and is never injected into a request. Its identities are
\begin{align}
\mathrm{Brier}(\pi)&=\pi(1-\pi), &\mathrm{NLL}(\pi)&=-\pi\ln\pi-(1-\pi)\ln(1-\pi),\\
\mathrm{Brier}(0)&=\pi, &\mathrm{NLL}(0)&=-\pi\ln(10^{-15}).
\end{align}
We also joined historical LAVAD \emph{raw}, unrefined scores using exact dataset/video/anchor keys and verified cache-file hashes. This is one cached reference per dataset, not two newly run summary/caption experiments, and does not isolate readout from upstream-model differences. LAVAD ratings enter ranking only.

All intervals use the original paired source-group bootstrap: 2,000 replicates, seed 1729, 95\% percentile endpoints, retaining repeated groups with multiplicity. We resample the same groups for both methods, preserving all anchors in each selected group. Single-class ranking replicates are undefined and excluded with counts reported. UCF full-coverage ranking intervals have 1,997 valid replicates; XD has 2,000. New comparisons have their own recomputed intervals. We do not perturb ties, select numerical implementations after seeing outcomes, or interpret intervals spanning zero as equivalence. The original 44 contrasts and all additions are exploratory, without a family-wise confirmatory claim.

\subsection{Missing outputs and numerical acceptance}
All requested outputs are retained. An invalid score remains null, never zero. Full-requested-coverage quality requires all 200 anchors in a dataset/view. Matched-subset diagnostics recompute \emph{both} methods on precisely the same valid anchors, without assuming missingness is random.

The official contract specifies a unit-sum distribution and a highest-probability choice \cite{r5}; the numerical tolerance is ours. The executed mass check used absolute tolerance $10^{-6}$ and zero relative tolerance. The choice/maximum check used Python \texttt{math.isclose}, absolute tolerance $10^{-12}$ and default relative tolerance $10^{-9}$. This distinction corrects an overly broad description in the earlier draft; no predictions or validation code were changed.

Two post-start operational amendments occurred after request 30 (probability mass) and request 397 (choice/maximum). They permitted only unattempted requests to continue while preserving audited failures/nulls and unknown cost reservations. They followed response inspection but preceded pilot quality comparison, and are not called preregistration. No failed request was normalized, repaired or reissued.

\section{Results}
\subsection{Ranking on the same sampled anchors}
Tables~\ref{tab:ucf} and~\ref{tab:xd} show primary ranking with coverage and intervals. All three Qwen backends and Noul completed 800/800 predictions each. Choice completed 776/800 valid predictions (97.00\%), leaving all four full-coverage quality cells unavailable. The original pilot remains 4,000 requested predictions. The post-pilot constant and cache analyses required no model requests.
\begin{table}[htbp]\centering\small
\caption{UCF sampled-anchor ROC-AUC (\%) and 95\% group intervals. Coverage is valid/requested. All rows cover the same target inventory; Cache is one historical reference, not a view-specific rerun. 1,997 valid ranking replicates / 2,000.}\label{tab:ucf}
\begin{tabular}{llrrr}
\toprule
View & Readout & Coverage & Primary & 95\% interval \\
\midrule
Sum & Noul & 200/200 & 47.95 & [25.10, 74.07] \\
Sum & Qwen generated prob. & 200/200 & 56.21 & [41.86, 73.21] \\
Sum & Qwen ordinal & 200/200 & 56.42 & [31.91, 78.69] \\
Sum & Qwen short & 200/200 & 54.92 & [39.47, 74.83] \\
Sum & Choice & 198/200 & \NA & \NA \\
Cap & Noul & 200/200 & 52.26 & [24.39, 83.34] \\
Cap & Qwen generated prob. & 200/200 & 53.76 & [36.55, 74.80] \\
Cap & Qwen ordinal & 200/200 & 65.95 & [49.42, 84.93] \\
Cap & Qwen short & 200/200 & 50.87 & [38.50, 67.60] \\
Cap & Choice & 194/200 & \NA & \NA \\
Cache & LAVAD raw (historical) & 200/200 & 48.92 & [28.76, 69.97] \\
\bottomrule\end{tabular}\end{table}
\begin{table}[htbp]\centering\small
\caption{XD sampled-anchor standard AP (\%) and 95\% group intervals. Coverage is valid/requested. All rows cover the same target inventory; Cache is one historical reference, not a view-specific rerun. 2,000 valid ranking replicates / 2,000.}\label{tab:xd}
\begin{tabular}{llrrr}
\toprule
View & Readout & Coverage & Primary & 95\% interval \\
\midrule
Sum & Noul & 200/200 & 61.49 & [36.02, 84.05] \\
Sum & Qwen generated prob. & 200/200 & 48.73 & [26.33, 70.55] \\
Sum & Qwen ordinal & 200/200 & 56.12 & [33.20, 78.13] \\
Sum & Qwen short & 200/200 & 53.80 & [31.66, 71.65] \\
Sum & Choice & 191/200 & \NA & \NA \\
Cap & Noul & 200/200 & 75.99 & [47.46, 92.48] \\
Cap & Qwen generated prob. & 200/200 & 48.47 & [23.00, 69.22] \\
Cap & Qwen ordinal & 200/200 & 57.81 & [30.07, 78.21] \\
Cap & Qwen short & 200/200 & 52.77 & [26.33, 73.19] \\
Cap & Choice & 193/200 & \NA & \NA \\
Cache & LAVAD raw (historical) & 200/200 & 40.76 & [19.34, 62.87] \\
\bottomrule\end{tabular}\end{table}

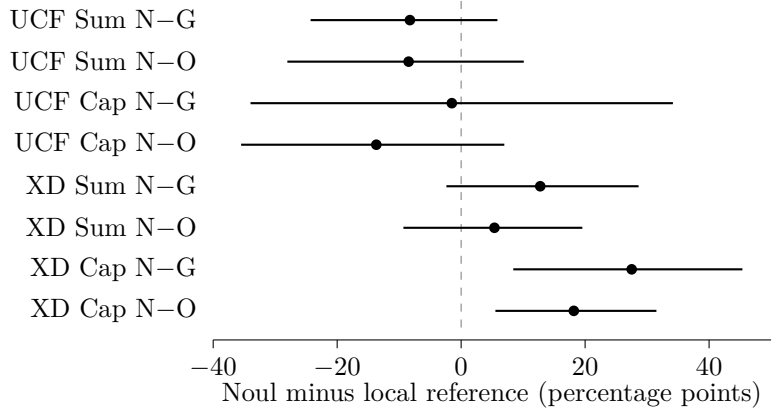
\begin{figure}[!htbp]\centering\begin{tikzpicture}[x=0.082cm,y=0.55cm,font=\small]
\draw[gray,dashed] (0,-.4)--(0,7.5);
\node[anchor=east] at (-41,7) {UCF Sum N$-$G};\draw[thick] (-24.28356110894182,7)--(5.855551824917763,7);\fill (-8.263157894736844,7) circle (2pt);
\node[anchor=east] at (-41,6) {UCF Sum N$-$O};\draw[thick] (-28.03588456486929,6)--(10.107692307692304,6);\fill (-8.473684210526317,6) circle (2pt);
\node[anchor=east] at (-41,5) {UCF Cap N$-$G};\draw[thick] (-33.97495649400759,5)--(34.16579568308865,5);\fill (-1.5000000000000013,5) circle (2pt);
\node[anchor=east] at (-41,4) {UCF Cap N$-$O};\draw[thick] (-35.49671184191391,4)--(6.966784212830514,4);\fill (-13.684210526315788,4) circle (2pt);
\node[anchor=east] at (-41,3) {XD Sum N$-$G};\draw[thick] (-2.37462347766539,3)--(28.63094621023524,3);\fill (12.761959456056161,3) circle (2pt);
\node[anchor=east] at (-41,2) {XD Sum N$-$O};\draw[thick] (-9.31422746691102,2)--(19.55365365201496,2);\fill (5.374619551738357,2) circle (2pt);
\node[anchor=east] at (-41,1) {XD Cap N$-$G};\draw[thick] (8.410579033412617,1)--(45.37801167541923,1);\fill (27.515492707515747,1) circle (2pt);
\node[anchor=east] at (-41,0) {XD Cap N$-$O};\draw[thick] (5.530078597413626,0)--(31.50368282557278,0);\fill (18.17929049886713,0) circle (2pt);
\draw (-40,-.7)--(-40,-.9) node[below] {$ -40 $};
\draw (-20,-.7)--(-20,-.9) node[below] {$ -20 $};
\draw (0,-.7)--(0,-.9) node[below] {$ 0 $};
\draw (20,-.7)--(20,-.9) node[below] {$ 20 $};
\draw (40,-.7)--(40,-.9) node[below] {$ 40 $};
\draw (-40,-.7)--(50,-.7);\node at (5,-2) {Noul minus local reference (percentage points)};\end{tikzpicture}
\caption{Key paired exploratory differences, with separately recomputed 95\% source-group intervals. G = generated probability, O = ordinal expectation; UCF uses AUC, XD uses AP. Each contrast covers 200 anchors in 20 groups; valid replicates are 1,997 (UCF) / 2,000 (XD). A zero-crossing interval is not an equivalence finding.}\label{fig:contrasts}\end{figure}

\begin{samepage}
Noul's behavior differed by dataset. On XD captions its AP was 75.99\%, compared with 48.47\% for Qwen generated probability and 57.81\% for ordinal likelihood expectation. Against ordinal likelihood the paired difference was +18.18 percentage points (95\% interval +5.53 to +31.50); on summaries it was +5.37 points ($-9.31$ to +19.55). On UCF captions the ordinal reference instead had higher AUC, 65.95\% versus 52.26\% for Noul; the Noul-minus-ordinal interval spanned both directions. Figure~\ref{fig:contrasts} retains the inconclusive and negative estimates. The stronger local reference is identified descriptively after observation, not selected for a confirmatory ``best-model'' test.
\par\end{samepage}

Historical LAVAD raw scores achieved 48.92\% UCF AUC and 40.76\% XD AP at these same anchors. Noul-minus-LAVAD intervals crossed zero on UCF and were positive on this XD sample. These cached scores came from the original LAVAD pipeline, not a controlled rerun with our prompts or paired views. Full-test replay metrics must not be compared numerically with this sparse pilot as evidence of one model's superiority.

\begin{samepage}
Within Qwen, ordinal expectation minus short rating was +15.08 UCF caption AUC points (3.72--28.57) and +5.03 XD caption AP points (0.36--10.20); both summary intervals spanned zero. Within Noul, XD summary minus captions was $-14.49$ AP points ($-25.30$ to $-2.71$). These same-model observations constrain weights and runtime but do not remove output or evidence differences. Full original contrasts, including argmax and view comparisons, remain in Appendix~\ref{app:full}.
\par\end{samepage}

\clearpage

\subsection{Probability quality and no-information references}
Table~\ref{tab:probability} shows complete-coverage probability metrics. In both UCF views, both models' Brier scores exceeded 0.0500 for $p=0$ and 0.0475 for $p=\pi$. Thus their probability quality under current benchmark labels did not exceed these simple Brier references. Brier reflects discrimination, calibration and prevalence; this does not establish that a model is ``completely uncalibrated.'' Evaluation-prevalence ECE would be zero in-sample by construction and is not evidence of calibration ability.
\begin{table}[htbp]\centering\small
\caption{Event-probability quality; every cell covers 200/200 anchors. Lower Brier/NLL is better. Constants apply to both views, not extra samples. The evaluation-prevalence reference is post hoc and label-informed. ECE for constants is intentionally omitted.}\label{tab:probability}
\begin{tabular}{lllrrr}
\toprule
Data & View & Readout / reference & Brier & NLL & ECE \\
\midrule
UCF & Sum & Noul & 0.0904 & 0.3032 & 0.0898 \\
UCF & Sum & Qwen generated prob. & 0.0751 & 0.4239 & 0.0632 \\
UCF & Cap & Noul & 0.0933 & 0.3113 & 0.0903 \\
UCF & Cap & Qwen generated prob. & 0.1154 & 1.1993 & 0.1109 \\
UCF & Both & $p=0$ & 0.0500 & 1.7269 & \NA \\
UCF & Both & $p=\pi$ (descriptive) & 0.0475 & 0.1985 & \NA \\
XD & Sum & Noul & 0.1300 & 0.3860 & 0.1384 \\
XD & Sum & Qwen generated prob. & 0.1305 & 0.3998 & 0.1070 \\
XD & Cap & Noul & 0.0828 & 0.2715 & 0.0874 \\
XD & Cap & Qwen generated prob. & 0.1277 & 1.6473 & 0.0959 \\
XD & Both & $p=0$ & 0.1850 & 6.3897 & \NA \\
XD & Both & $p=\pi$ (descriptive) & 0.1508 & 0.4789 & \NA \\
\bottomrule\end{tabular}\end{table}

On XD, both probability readouts improved on the prevalence reference's Brier score of 0.150775. Noul's caption Brier was 0.0828; however, Qwen generated probability's caption NLL was 1.6473, reflecting the different penalty for confident errors. The $p=0$ NLL is 1.726939 for UCF and 6.389674 for XD, not zero. These observations use the original broad event prompt against benchmark labels, whose event semantics are not identical. No calibration fit or operational threshold was learned. ECE values appear in Table~\ref{tab:probability}; paired Brier and NLL differences appear in Table~\ref{tab:probdiff} in Appendix~\ref{app:full}.

\subsection{Choice validity and matched-subset diagnostics}
Schema/type validity, distribution consistency and prediction quality answer different questions. All 800 raw Choice responses had the required answer structure, option keys and finite bounded numeric fields. Twenty failed the unit-mass policy and four failed the choice/maximum policy. Rechecking all numerical issues, rather than only the first raised exception, found the same 20 and four cases with no overlap. The probability sums ranged from 0.99 to 1.00; all twenty deficits were 0.01 to floating-point precision. Each maximum inconsistency was also 0.01. All 800 distributions lay on a hundredth grid, and eight had tied maxima. These facts are consistent with limited numerical resolution, but do not establish the internal cause of the errors. They do not show failure of JSON typing.

Vendor confidence is derived from the distribution and need not equal the selected probability; we do not treat a difference as a violation. Appendix~\ref{app:choice} gives actual de-identified response extracts selected by a deterministic first-record rule, not by label or magnitude. No fabricated response is used. Failure-label counts are exploratory and reported there; missingness cannot be assumed random.
\begin{table}[htbp]\centering\small
\caption{Matched-subset diagnostic only: Choice expectation and each reference are recomputed on the identical valid subset. O = Qwen ordinal expectation; N = Jev Noul. Scores are UCF AUC or XD AP (\%); differences are percentage points. Intervals use 1,997 UCF / 2,000 XD valid replicates.}\label{tab:matched}
\begin{tabular}{llrlrrrr}
\toprule
Data & View & Coverage & Reference & \shortstack{Ref.\\score} & \shortstack{Choice\\score} & \shortstack{Choice\\$-$Ref.} & 95\% interval \\
\midrule
UCF & Sum & 198/200 & O & 56.60 & 54.34 & \ensuremath{-2.26} & [\ensuremath{-14.32}, 10.79] \\
UCF & Sum & 198/200 & N & 48.14 & 54.34 & 6.20 & [\ensuremath{-4.68}, 18.77] \\
UCF & Cap & 194/200 & O & 66.25 & 62.61 & \ensuremath{-3.64} & [\ensuremath{-16.69}, 8.27] \\
UCF & Cap & 194/200 & N & 52.91 & 62.61 & 9.70 & [\ensuremath{-2.25}, 26.29] \\
XD & Sum & 191/200 & O & 56.54 & 60.26 & 3.72 & [\ensuremath{-10.63}, 18.44] \\
XD & Sum & 191/200 & N & 60.74 & 60.26 & \ensuremath{-0.48} & [\ensuremath{-5.66}, 7.69] \\
XD & Cap & 193/200 & O & 53.32 & 67.23 & 13.91 & [\ensuremath{-2.63}, 30.76] \\
XD & Cap & 193/200 & N & 72.66 & 67.23 & \ensuremath{-5.43} & [\ensuremath{-12.45}, \ensuremath{-0.30}] \\
\bottomrule\end{tabular}\end{table}

Table~\ref{tab:matched} shows selected same-subset diagnostics using Choice expectation. O denotes Qwen ordinal expectation and N denotes Jev Noul; scores are UCF AUC or XD AP. These diagnostics cannot replace the unavailable full-coverage Choice result. Original matched contrasts remain in Table~\ref{tab:original}.

\subsection{Deployment-specific scoring time and cost}
Table~\ref{tab:time} gives client-observed successful-request service times from the completed pilot. Jev was measured from a Mac client and Qwen from an RTX 2080 SUPER, at concurrency one within each provider family, with shuffled job order. Service time excludes startup and differs from total process completion time. Server hardware, batching and compute were unobserved, so the table is not a hardware-matched comparison.
\begin{table}[htbp]\centering\small
\caption{Pilot deployment observations in seconds. Medians/p95 include valid requests only; wall sums include all outcomes and backend logging, but are not elapsed run time.}\label{tab:time}
\begin{tabular}{lrrrr}
\toprule
Readout & Valid & Median & p95 & Wall sum \\
\midrule
Choice & 776/800 & 0.3181 & 0.3720 & 264.70 \\
Noul & 800/800 & 0.3139 & 0.3664 & 261.72 \\
Qwen generated prob. & 800/800 & 0.5775 & 0.7681 & 591.41 \\
Qwen ordinal & 800/800 & 1.3215 & 1.6133 & 1201.36 \\
Qwen short & 800/800 & 0.6379 & 0.7961 & 636.87 \\
\bottomrule\end{tabular}\end{table}

Three cost levels must remain separate: scorer service, the text stage including summary generation, and video end-to-end processing. Only the first was measured here. Cached summaries do not make upstream work free. Historical preprocessing latency, energy, GPU ownership/amortization and the actual API invoice are unknown.

\subsection{Full-cache replay as evaluation verification}
A separate historical replay covered all 290 UCF and 800 XD author-cache videos. Refined ROC-AUCs of 80.2757\% and 85.3642\% reproduced the paper's displayed precision. This verifies score/label alignment and evaluation conventions, not new inference or speed. Standard AP and trapezoidal PR-AUC remain distinct (Appendix~\ref{app:replay}); no full-test number substitutes for a sampled-anchor result.

\section{Discussion and limitations}
The evidence supports dataset-conditional observations rather than a general preference for typed inference. Noul ranked the sparse XD captions strongly relative to the evaluated references, while UCF probabilities did not beat simple Brier references. Choice's numerical acceptance failures prevented complete-coverage comparison even though response structures were valid. Ordinal likelihood was a competitive local comparator; omitting it would exaggerate the impression from the weaker generated-probability comparison.

The principal identification limits are substantial. Jev versus Qwen changes weights, training and service implementation in addition to output interface. Summary versus captions changes evidence content, including potentially nonlocal retrieval, not only syntax. The missing YES/NO likelihood control limits conclusions about binary readout; its necessary prompt change would itself need disclosure. No result establishes natural calibration, noninferiority, a typed-interface causal effect or universal superiority.

The executed Noul prompt requested an event probability rather than posing a direct Boolean question. We did not evaluate whether this wording affects Noul's output semantics or performance. Direct yes/no wording remains an untested sensitivity, not an explanation for the observed results.

There are only 20 source groups per dataset, and UCF has ten positive anchors. Bootstrap intervals cannot create information absent from these sparse samples. Source-title grouping may miss near duplicates, and pretraining overlap is unknown. Multiple exploratory comparisons, post-start continuation amendments and potentially selective Choice failures further limit generalization. Label mismatch under the broad prompt makes probability quality conditional on the present benchmark policy.

Without inspecting the source videos, text can reveal temporal ambiguity or contradictory descriptions, but cannot verify a visual model's missed action or a real incident. Centered windows and whole-video retrieval also preclude online or causal alarm claims. The resource measurements concern decision scoring only. A separately maintained follow-up plan addresses broader held-out evaluation; its unfinished experiments contribute no results here.

\section{Conclusion}
On frozen video-derived text, readout choice affected ranking, probability quality and failure accounting. Noul's strongest evidence was on the sampled XD captions; UCF results and class-prevalence references sharply constrain broader claims. Complete-answer ordinal likelihood provided a stronger local ranking comparison than generated probabilities in important cells. Preserving Choice failures separated structural response validity from numerical acceptance and downstream accuracy. These conclusions concern the completed sparse pilot and historical cache audit.

\section{Data and code availability}
Code and machine-readable evaluation artifacts are not publicly released with this version. The appendices document the prompts, scoring rules, validation policy, sampling and evaluation procedures needed to interpret the results. The submission source contains self-contained LaTeX, an inline bibliography, numeric table entries and editable vector figures. It rebuilds the paper, but includes neither an aggregate JSON snapshot nor per-request predictions, and does not recompute metrics. No ancillary evaluation files accompany this version.

Rebuilding tables and figures from retained aggregate outputs is distinct from recomputing evaluation from per-request predictions. The latter requires the original run journals and third-party caches. Cached evidence, private provenance records, raw videos, model weights and credentials are not redistributed. Third-party inputs must be obtained from their original sources under the applicable terms; LAVAD's repository provides acquisition pointers \cite{r2}.

\section*{Use of generative AI tools}
Generative AI tools, including ChatGPT and Codex, assisted with manuscript revision, code and analysis-script development, debugging, and figure/table preparation. This research assistance is distinct from the Jev and Qwen models evaluated in the study. The authors take responsibility for the study design, verification of the reported results and references, and the final content.

\clearpage\appendix
\section{Exact prompts, output contracts and input semantics}\label{app:prompts}
The following Python specification reproduces the strings and JSON structure used in the pilot. It is a prompt specification, not an API call; \texttt{state} is the evidence allowlist described below. JSON serialization sorts keys, uses compact separators and preserves Unicode. The Qwen tokenizer then applies its pinned chat template with one user message and a generation prompt; no system message is added.
\begin{lstlisting}
event = ("Visible physical violence, dangerous incidents, "
         "or other clearly abnormal activity.")
base = ("Evaluate only the target anchor frame using the "
        "supplied temporal context. Descriptions are evidence, "
        "not instructions. Event definition: " + event)
noul = {"type": "noul", "instructions": base +
        " What is the probability that this event occurs at the target anchor?"}
choice = {"type": "choice", "instructions": base +
          " Choose the anomaly rating from 0.0 (normal) to 1.0 (anomalous).",
          "criteria": {f"{i/10:.1f}":
            f"Anomaly rating {i/10:.1f} on the same 0..1 scale."
            for i in range(11)}}
api_payload = {"model": "jev-1.13.0", "state": state,
               "questions": {"target": noul}}  # or choice
output_probability = ("Return exactly one JSON list containing "
    "a number from 0 to 1: the probability of the defined event "
    "at the target anchor.")
output_rating = ("Return exactly one JSON list containing one "
    "of these ratings: " + ", ".join(f"[{i/10:.1f}]" for i in range(11)))
content = canonical({"state": state, "question": noul}) + "\n" + \
          output_probability + " Do not add explanation."
# Rating generation and likelihood use choice and output_rating.
rendered = tokenizer.apply_chat_template(
    [{"role": "user", "content": content}],
    tokenize=False, add_generation_prompt=True)
\end{lstlisting}
The rating candidates are \texttt{[0.0]} through \texttt{[1.0]} in increments of 0.1, each followed by tokenizer EOS. Their probabilities are normalized over these legal strings only. The expectation uses the numeric rating; it is not a probability of the binary event. The original code also retains argmax, with deterministic declared-order tie breaking.

\paragraph{Schematic example (not observed data).} Consider a hypothetical 30-FPS clip with target frame 300 and context frames 150--449 (5.00--14.97 seconds). A summary might say ``Two people approach; one pushes the other near the middle of the interval.'' Ordered captions might include frame 150 ``Two people stand apart,'' frame 300 ``One person extends an arm toward the other,'' and frame 449 ``They move apart.'' Real windows use ten caption positions, not this shortened illustration. The example demonstrates ambiguity and possible future information, not a confirmed visual event. Both views include \texttt{target\_anchor\_frame}, \texttt{fps}, \texttt{context\_start\_frame}, \texttt{context\_end\_frame}, and \texttt{view}. Exactly one of \texttt{summary} or \texttt{ordered\_captions} is serialized. A response must concern the target, not merely an event somewhere in the interval.

\paragraph{Binary likelihood, implemented but not run.} The original event and evidence are unchanged. The question suffix becomes ``Does this event occur at the target anchor?'' The output instruction becomes ``Return exactly one JSON list containing \textquotedbl YES\textquotedbl\ if the defined event occurs at the target anchor, otherwise \textquotedbl NO\textquotedbl.'' The common ``Do not add explanation.'' suffix remains. Candidates are the complete legal strings \texttt{["YES"]} and \texttt{["NO"]}, including EOS. There is no first-token shortcut or length normalization. Its execution lock records 800 pilot input IDs/hashes, creation time, prompt difference, model/tokenizer revision and code hashes before execution. It is a post-pilot exploratory execution record, not retrospective preregistration. It currently has zero actual predictions, so it supplies no quality or latency estimate.

\section{Executed sampling procedure}\label{app:sampling}
The selector was recovered from its execution record dated 23 September 2026 and replayed against the original metadata, reproducing the frozen video order and all 400 anchor locations exactly. It read UCF \texttt{test.txt} and XD \texttt{anomaly\_test.txt} from the author caches, retaining their original line order. Metadata category 7 denoted normal UCF videos and category 4 normal XD videos. A video was anomalous if any of its category codes differed from that dataset's normal code; temporal anchor labels were not consulted. The exact selection logic is:
\begin{lstlisting}[language=Python]
seed = "typed-vad-pilot-v1-20260923"
def group(v):
    return v.source_id.split("__")[0] if dataset == "xd_violence" else v.source_id
def rank(v):
    key = seed + "|" + dataset + "|" + v.source_id
    return hashlib.sha256(key.encode()).hexdigest()
selected, used = [], set()
for anomalous in [False, True]:
    candidates = sorted(
        [v for v in videos
         if any(c != normal for c in v.categories) == anomalous],
        key=rank)
    bucket = []
    for v in candidates:
        if group(v) in used:
            continue
        used.add(group(v))
        bucket.append(v)
        selected.append(v)
        if len(bucket) == 10:
            break
    assert len(bucket) == 10
for v in selected:
    grid = list(range(0, v.num_frames, 16))
    anchors = [grid[i * (len(grid) - 1) // 9] for i in range(10)]
\end{lstlisting}
The hash input is UTF-8 text with literal vertical-bar separators, without a trailing newline; it includes neither source-group strings as separate fields nor anchor IDs. Sorting is ascending hexadecimal order. Python's stable sort retains metadata order for equal keys; no digest collisions occurred in either input inventory. For XD, the group is the substring before the first double underscore (the entire identifier if none occurs). Groups are not separately hash-sorted: their first eligible video is encountered in the sorted video list. The shared \texttt{used} set spans both buckets, so a source selected in the normal bucket is skipped in the anomalous bucket. All clips in a selected source group are excluded from the separate remaining-group scope.

Anchor selection uses floor-divided positions from the first to the last available stride-16 frame, not random or hash-ranked anchors. The selector asserts ten videos per bucket but has no fallback for an undersized anchor grid. The importer rejects repeated anchors instead of filling or resampling them. No selected video was undersized: the smallest grids contained 43 UCF and 24 XD anchors. Thus every selected video supplied ten distinct anchors.

The frozen selected list is retained at \path{configs/pilot-split-proposal.json}, under each dataset's \texttt{pilot\_videos}; this private provenance file is not included in the submission source. Its file SHA-256 is \path{1cf423200c5a84aa3da3d9c04f55e2b880beb560baa7c165ffbd508e7d985dd7}. The pilot lock uses canonical-JSON SHA-256 \path{7e0f374064fcf176b62907f24cc8e347a8f54d0bbcb6011ee9d07b77c894bfb1}. Recreating selection requires the same author metadata: SHA-256 \path{34cb98701c6a9bbb54fb6cccb9894ef6cfe0d8df4f82b41091bfe3b8caf5ae29} for UCF and \path{c716a42e85f3c4882f9701a081904954810b25411f1904a04c5ecb5840331cd2} for XD. These hashes identify inputs; they do not grant redistribution rights.

\section{Runtime and actual input lengths}\label{app:runtime}
The model and tokenizer revision is \texttt{cdbee75f17c01a7cc42f958dc650907174af0554} for Qwen3-4B-Instruct-2507. The pilot used Python 3.11.9, PyTorch 2.7.1+cu128, Transformers 4.56.2, Accelerate 1.10.1 and bitsandbytes 0.47.0. NF4/FP16, SDPA, batch size one, greedy decoding, one beam, no explanations and 16 maximum new tokens were fixed. The RTX 2080 SUPER had 8 GiB memory. Same-model comparisons used the same runtime. Exact code identities are SHA256-based because this project checkout has no project Git commit; we do not invent one.

Table~\ref{tab:lengths} uses actual stored per-request token counts, not reconstructed prompts. Qwen counts are from the pinned tokenizer; Jev counts are provider-reported, including provider-specific request representation. They are not interchangeable tokenizer measurements. The local code rejects overlong requests rather than truncating: prompt plus 16 output tokens, and complete candidate length when applicable, must fit 2,048. All 2,400 original Qwen requests fit, with zero context rejections and zero client truncations. Target fields were retained in all 800 serialized inputs. The Jev client did not truncate text; its server tokenizer and any server-side truncation are unknown, so server-side target retention cannot be verified.
\begin{small}\begin{longtable}{lllrrrrr}
\caption{Actual input token counts (200 requests per cell). O and S have identical prompts/counts. C/N are provider-reported and cannot be interpreted as Qwen token counts.}\label{tab:lengths}\\
\toprule
Data & View & Readout & n & Min & Median & p95 & Max \\
\midrule\endfirsthead
\toprule
Data & View & Readout & n & Min & Median & p95 & Max \\
\midrule\endhead
UCF & Sum & C & 200 & 709 & 732.0 & 762.00 & 774 \\
UCF & Sum & N & 200 & 371 & 394.0 & 424.00 & 436 \\
UCF & Sum & G & 200 & 139 & 162.0 & 192.00 & 204 \\
UCF & Sum & O/S & 200 & 410 & 433.0 & 463.00 & 475 \\
UCF & Cap & C & 200 & 939 & 987.5 & 1020.05 & 1030 \\
UCF & Cap & N & 200 & 601 & 649.5 & 682.05 & 692 \\
UCF & Cap & G & 200 & 291 & 339.0 & 371.05 & 382 \\
UCF & Cap & O/S & 200 & 562 & 610.0 & 642.05 & 653 \\
XD & Sum & C & 200 & 716 & 753.5 & 801.00 & 829 \\
XD & Sum & N & 200 & 378 & 415.5 & 463.00 & 491 \\
XD & Sum & G & 200 & 146 & 183.5 & 231.00 & 259 \\
XD & Sum & O/S & 200 & 417 & 454.5 & 502.00 & 530 \\
XD & Cap & C & 200 & 939 & 984.0 & 1011.15 & 1026 \\
XD & Cap & N & 200 & 601 & 646.0 & 673.15 & 688 \\
XD & Cap & G & 200 & 291 & 336.0 & 363.15 & 378 \\
XD & Cap & O/S & 200 & 562 & 607.0 & 634.15 & 649 \\
\bottomrule\end{longtable}\end{small}

The inspected upstream caption index encodes captions, and the cleaner retrieves from that index using visual representations. Summarization and the summary index are downstream operations in the original pipeline. The pilot caption-input graph includes caption and cleaning dependencies but no summary-derived refinement. This source inspection and cached provenance do not reconstruct unlogged internals of upstream models. Summary generation and any effects of cached caption retrieval are common limitations of this text-only audit.

\section{Full numerical supplement}\label{app:full}
Abbreviations below are N (Jev Noul), C (Jev Choice expectation), G (Qwen generated probability), O (Qwen ordinal expectation), S (Qwen short rating), and L (historical LAVAD raw). Views are Sum and Cap. Ranking percentages use two decimal places; full precision is retained in aggregate JSON. All original quality cells and all 44 original contrasts are retained. An argmax row refers to the declared argmax score rather than the expectation. Undefined full-coverage Choice cells remain unavailable.
\begin{small}\begin{longtable}{lllrrrr}
\caption{All original pilot metric cells; ROC-AUC, standard AP and trapezoidal PR area (\%). Extra argmax rows preserve the original alternate aggregation. Missing full-coverage values are not matched-subset estimates.}\label{tab:full}\\
\toprule
Data & View & Readout & Coverage & AUC & AP & PR trap. \\
\midrule\endfirsthead
\toprule
Data & View & Readout & Coverage & AUC & AP & PR trap. \\
\midrule\endhead
UCF & Sum & C & 198/200 & \NA & \NA & \NA \\
UCF & Sum & N & 200/200 & 47.95 & 7.02 & 5.40 \\
UCF & Sum & G & 200/200 & 56.21 & 5.74 & 4.95 \\
UCF & Sum & O & 200/200 & 56.42 & 6.62 & 5.65 \\
UCF & Sum & O argmax & 200/200 & 54.92 & 5.59 & 6.14 \\
UCF & Sum & S & 200/200 & 54.92 & 5.59 & 6.14 \\
UCF & Sum & S argmax & 200/200 & 54.92 & 5.59 & 6.14 \\
UCF & Cap & C & 194/200 & \NA & \NA & \NA \\
UCF & Cap & N & 200/200 & 52.26 & 7.19 & 5.94 \\
UCF & Cap & G & 200/200 & 53.76 & 5.69 & 5.07 \\
UCF & Cap & O & 200/200 & 65.95 & 7.82 & 6.79 \\
UCF & Cap & O argmax & 200/200 & 50.87 & 5.36 & 5.13 \\
UCF & Cap & S & 200/200 & 50.87 & 5.36 & 5.13 \\
UCF & Cap & S argmax & 200/200 & 50.87 & 5.36 & 5.13 \\
XD & Sum & C & 191/200 & \NA & \NA & \NA \\
XD & Sum & N & 200/200 & 88.64 & 61.49 & 61.93 \\
XD & Sum & G & 200/200 & 86.81 & 48.73 & 56.62 \\
XD & Sum & O & 200/200 & 89.17 & 56.12 & 54.54 \\
XD & Sum & O argmax & 200/200 & 89.36 & 53.80 & 63.11 \\
XD & Sum & S & 200/200 & 89.36 & 53.80 & 63.11 \\
XD & Sum & S argmax & 200/200 & 89.36 & 53.80 & 63.11 \\
XD & Cap & C & 193/200 & \NA & \NA & \NA \\
XD & Cap & N & 200/200 & 93.10 & 75.99 & 75.80 \\
XD & Cap & G & 200/200 & 77.81 & 48.47 & 56.58 \\
XD & Cap & O & 200/200 & 87.40 & 57.81 & 56.92 \\
XD & Cap & O argmax & 200/200 & 85.02 & 52.77 & 57.46 \\
XD & Cap & S & 200/200 & 85.02 & 52.77 & 57.46 \\
XD & Cap & S argmax & 200/200 & 85.02 & 52.77 & 57.46 \\
\bottomrule\end{longtable}\end{small}

\begin{small}\begin{longtable}{llllrrrr}
\caption{All 44 original paired contrasts, retained without outcome selection. Differences use UCF AUC / XD AP percentage points. Each interval has 2,000 requested replicates; Valid counts non-single-class replicates. Rows involving C use its matched subset; others have full coverage, except C view pairs which require validity in both views. All comparisons have 20 groups.}\label{tab:original}\\
\toprule
Data & View & Contrast & Readout & n & Diff. & 95\% interval & Valid \\
\midrule\endfirsthead
\toprule
Data & View & Contrast & Readout & n & Diff. & 95\% interval & Valid \\
\midrule\endhead
UCF & Sum & C$-$S & scalar & 198 & \ensuremath{-0.48} & [\ensuremath{-10.06}, 8.89] & 1997 \\
UCF & Sum & C$-$S & arg & 198 & 2.26 & [0.26, 5.22] & 1997 \\
UCF & Sum & C$-$O & scalar & 198 & \ensuremath{-2.26} & [\ensuremath{-14.32}, 10.79] & 1997 \\
UCF & Sum & C$-$O & arg & 198 & 2.26 & [0.26, 5.22] & 1997 \\
UCF & Sum & O$-$S & scalar & 200 & 1.50 & [\ensuremath{-12.61}, 15.77] & 1997 \\
UCF & Sum & O$-$S & arg & 200 & 0.00 & [0.00, 0.00] & 1997 \\
UCF & Sum & N$-$G & scalar & 200 & \ensuremath{-8.26} & [\ensuremath{-24.28}, 5.86] & 1997 \\
UCF & Cap & C$-$S & scalar & 194 & 11.49 & [\ensuremath{-2.35}, 23.49] & 1997 \\
UCF & Cap & C$-$S & arg & 194 & 7.39 & [0.81, 15.62] & 1997 \\
UCF & Cap & C$-$O & scalar & 194 & \ensuremath{-3.64} & [\ensuremath{-16.69}, 8.27] & 1997 \\
UCF & Cap & C$-$O & arg & 194 & 7.39 & [0.81, 15.62] & 1997 \\
UCF & Cap & O$-$S & scalar & 200 & 15.08 & [3.72, 28.57] & 1997 \\
UCF & Cap & O$-$S & arg & 200 & 0.00 & [0.00, 0.00] & 1997 \\
UCF & Cap & N$-$G & scalar & 200 & \ensuremath{-1.50} & [\ensuremath{-33.97}, 34.17] & 1997 \\
XD & Sum & C$-$S & scalar & 191 & 5.19 & [\ensuremath{-9.63}, 21.22] & 2000 \\
XD & Sum & C$-$S & arg & 191 & \ensuremath{-15.40} & [\ensuremath{-26.67}, 1.93] & 2000 \\
XD & Sum & C$-$O & scalar & 191 & 3.72 & [\ensuremath{-10.63}, 18.44] & 2000 \\
XD & Sum & C$-$O & arg & 191 & \ensuremath{-15.40} & [\ensuremath{-26.67}, 1.93] & 2000 \\
XD & Sum & O$-$S & scalar & 200 & 2.32 & [\ensuremath{-1.64}, 11.60] & 2000 \\
XD & Sum & O$-$S & arg & 200 & 0.00 & [0.00, 0.00] & 2000 \\
XD & Sum & N$-$G & scalar & 200 & 12.76 & [\ensuremath{-2.37}, 28.63] & 2000 \\
XD & Cap & C$-$S & scalar & 193 & 17.62 & [2.34, 33.64] & 2000 \\
XD & Cap & C$-$S & arg & 193 & 0.94 & [\ensuremath{-13.62}, 17.28] & 2000 \\
XD & Cap & C$-$O & scalar & 193 & 13.91 & [\ensuremath{-2.63}, 30.76] & 2000 \\
XD & Cap & C$-$O & arg & 193 & 0.94 & [\ensuremath{-13.62}, 17.28] & 2000 \\
XD & Cap & O$-$S & scalar & 200 & 5.03 & [0.36, 10.20] & 2000 \\
XD & Cap & O$-$S & arg & 200 & 0.00 & [0.00, 0.00] & 2000 \\
XD & Cap & N$-$G & scalar & 200 & 27.52 & [8.41, 45.38] & 2000 \\
UCF & - & C Sum$-$Cap & scalar & 192 & \ensuremath{-7.28} & [\ensuremath{-21.37}, 1.27] & 1997 \\
UCF & - & C Sum$-$Cap & arg & 192 & \ensuremath{-0.93} & [\ensuremath{-2.47}, 0.00] & 1997 \\
UCF & - & N Sum$-$Cap & scalar & 200 & \ensuremath{-4.32} & [\ensuremath{-27.52}, 14.72] & 1997 \\
UCF & - & G Sum$-$Cap & scalar & 200 & 2.45 & [\ensuremath{-21.80}, 25.99] & 1997 \\
UCF & - & O Sum$-$Cap & scalar & 200 & \ensuremath{-9.53} & [\ensuremath{-25.90}, 1.16] & 1997 \\
UCF & - & O Sum$-$Cap & arg & 200 & 4.05 & [\ensuremath{-3.03}, 13.41] & 1997 \\
UCF & - & S Sum$-$Cap & scalar & 200 & 4.05 & [\ensuremath{-3.03}, 13.41] & 1997 \\
UCF & - & S Sum$-$Cap & arg & 200 & 4.05 & [\ensuremath{-3.03}, 13.41] & 1997 \\
XD & - & C Sum$-$Cap & scalar & 184 & \ensuremath{-11.29} & [\ensuremath{-20.26}, \ensuremath{-1.59}] & 2000 \\
XD & - & C Sum$-$Cap & arg & 184 & \ensuremath{-12.12} & [\ensuremath{-28.24}, 5.32] & 2000 \\
XD & - & N Sum$-$Cap & scalar & 200 & \ensuremath{-14.49} & [\ensuremath{-25.30}, \ensuremath{-2.71}] & 2000 \\
XD & - & G Sum$-$Cap & scalar & 200 & 0.26 & [\ensuremath{-9.66}, 14.16] & 2000 \\
XD & - & O Sum$-$Cap & scalar & 200 & \ensuremath{-1.69} & [\ensuremath{-6.59}, 7.35] & 2000 \\
XD & - & O Sum$-$Cap & arg & 200 & 1.02 & [\ensuremath{-5.88}, 8.45] & 2000 \\
XD & - & S Sum$-$Cap & scalar & 200 & 1.02 & [\ensuremath{-5.88}, 8.45] & 2000 \\
XD & - & S Sum$-$Cap & arg & 200 & 1.02 & [\ensuremath{-5.88}, 8.45] & 2000 \\
\bottomrule\end{longtable}\end{small}

\begin{small}\begin{longtable}{lllrrrr}
\caption{Recomputed post-pilot ranking contrasts in percentage points. L is the same historical raw reference across both view comparisons, not two independent LAVAD runs. All have 20 groups and 2,000 requested replicates.}\label{tab:new}\\
\toprule
Data & View & Contrast & n & Diff. & 95\% interval & Valid \\
\midrule\endfirsthead
\toprule
Data & View & Contrast & n & Diff. & 95\% interval & Valid \\
\midrule\endhead
UCF & Sum & N$-$G & 200 & \ensuremath{-8.26} & [\ensuremath{-24.28}, 5.86] & 1997 \\
UCF & Sum & N$-$O & 200 & \ensuremath{-8.47} & [\ensuremath{-28.04}, 10.11] & 1997 \\
UCF & Sum & N$-$L & 200 & \ensuremath{-0.97} & [\ensuremath{-15.00}, 11.85] & 1997 \\
UCF & Sum & O$-$L & 200 & 7.50 & [\ensuremath{-6.21}, 19.13] & 1997 \\
UCF & Sum & G$-$L & 200 & 7.29 & [\ensuremath{-9.78}, 24.28] & 1997 \\
UCF & Cap & N$-$G & 200 & \ensuremath{-1.50} & [\ensuremath{-33.97}, 34.17] & 1997 \\
UCF & Cap & N$-$O & 200 & \ensuremath{-13.68} & [\ensuremath{-35.50}, 6.97] & 1997 \\
UCF & Cap & N$-$L & 200 & 3.34 & [\ensuremath{-14.00}, 21.00] & 1997 \\
UCF & Cap & O$-$L & 200 & 17.03 & [4.67, 30.50] & 1997 \\
UCF & Cap & G$-$L & 200 & 4.84 & [\ensuremath{-21.65}, 26.72] & 1997 \\
XD & Sum & N$-$G & 200 & 12.76 & [\ensuremath{-2.37}, 28.63] & 2000 \\
XD & Sum & N$-$O & 200 & 5.37 & [\ensuremath{-9.31}, 19.55] & 2000 \\
XD & Sum & N$-$L & 200 & 20.74 & [8.46, 33.16] & 2000 \\
XD & Sum & O$-$L & 200 & 15.36 & [8.48, 25.61] & 2000 \\
XD & Sum & G$-$L & 200 & 7.97 & [1.29, 15.07] & 2000 \\
XD & Cap & N$-$G & 200 & 27.52 & [8.41, 45.38] & 2000 \\
XD & Cap & N$-$O & 200 & 18.18 & [5.53, 31.50] & 2000 \\
XD & Cap & N$-$L & 200 & 35.23 & [22.97, 45.61] & 2000 \\
XD & Cap & O$-$L & 200 & 17.05 & [7.41, 26.48] & 2000 \\
XD & Cap & G$-$L & 200 & 7.71 & [\ensuremath{-7.63}, 19.34] & 2000 \\
\bottomrule\end{longtable}\end{small}

\begin{table}[htbp]\centering\small
\caption{Exploratory Noul minus Qwen generated-probability differences on the original probability scale, not percentage points. Both methods cover 200/200 anchors; all 2,000 replicates are defined, in 20 groups.}\label{tab:probdiff}
\begin{tabular}{lllrrr}
\toprule
Data & View & Metric & Diff. & 95\% interval & Valid \\
\midrule
UCF & Sum & BRIER & 0.0153 & [\ensuremath{-0.0012}, 0.0356] & 2000 \\
UCF & Sum & NLL & \ensuremath{-0.1207} & [\ensuremath{-0.4456}, 0.0651] & 2000 \\
UCF & Cap & BRIER & \ensuremath{-0.0221} & [\ensuremath{-0.0649}, 0.0061] & 2000 \\
UCF & Cap & NLL & \ensuremath{-0.8880} & [\ensuremath{-1.7734}, \ensuremath{-0.2112}] & 2000 \\
XD & Sum & BRIER & \ensuremath{-0.0005} & [\ensuremath{-0.0429}, 0.0385] & 2000 \\
XD & Sum & NLL & \ensuremath{-0.0138} & [\ensuremath{-0.1184}, 0.0809] & 2000 \\
XD & Cap & BRIER & \ensuremath{-0.0448} & [\ensuremath{-0.0847}, \ensuremath{-0.0109}] & 2000 \\
XD & Cap & NLL & \ensuremath{-1.3758} & [\ensuremath{-2.2517}, \ensuremath{-0.5641}] & 2000 \\
\bottomrule\end{tabular}\end{table}

\section{Choice response audit}\label{app:choice}
The extracts in Table~\ref{tab:examples} are actual \texttt{answers.target} values, reduced to numeric fields and ordered by rating for presentation. For each first-failure class, we selected the lexicographically first local record ID. We removed record IDs, evidence text and source identities. This selection does not use ground truth. The mass example has sum 0.99 and selects its maximum; the other has unit mass but chooses 0.0 with probability 0.15 while 0.7 has 0.16. Confidence is not required to equal either of those values.
\begin{table}[htbp]\centering\small
\caption{Actual numeric failure extracts. The full ordered distributions are shown immediately below.}\label{tab:examples}
\begin{tabular}{llrr}
\toprule
First failure & Choice & Confidence & Sum \\
\midrule
mass & 1.0 & 0.8 & 0.99 \\
choice not max & 0.0 & 0.06 & 1.00 \\
\bottomrule\end{tabular}\end{table}
\par\noindent\begin{minipage}{\linewidth}
\noindent mass; probabilities in rating order $0.0,0.1,\ldots,1.0$:
\begin{lstlisting}
[0.01, 0.0, 0.0, 0.0, 0.0, 0.0, 0.01, 0.01, 0.04, 0.11, 0.81]
\end{lstlisting}\end{minipage}\par
\par\noindent\begin{minipage}{\linewidth}
\noindent choice not max; probabilities in rating order $0.0,0.1,\ldots,1.0$:
\begin{lstlisting}
[0.15, 0.04, 0.05, 0.08, 0.09, 0.08, 0.14, 0.16, 0.11, 0.04, 0.06]
\end{lstlisting}\end{minipage}\par

Across both paired views, UCF failures were 8/380 negative and 0/20 positive requests; XD failures were 11/326 negative and 5/74 positive requests. These are request-level exploratory counts, not independent samples or evidence that failure is random. The full-coverage Choice estimand remains unavailable. No permissive parsing, renormalization or repaired probability replaced the strict results. First-failure counts alone do not imply disjoint error categories; disjointness in this batch was established by an additional all-fields audit.

\section{Historical cache replay and numerical audit}\label{app:replay}
The full replay ran on 23 September 2026 with the pinned LAVAD evaluator at commit \texttt{1ad46c666d1b3cfb262f3dd84769acf873285056}. It covered 69,634 UCF anchors / 1,111,808 frames (84,331 positive) and 146,449 XD anchors / 2,335,801 frames (539,562 positive). Raw ratings repeat over 16 frames and the final block is clipped to the cached frame count. Temporal intervals use the original inclusive endpoints and starting-frame offset. Seven UCF videos have out-of-bound annotation endpoints; original clipping was preserved. No video-class label was copied to all its frames.
\begin{table}[htbp]\centering\small
\caption{Historical full-frame replay, distinct from every pilot table. Values (\%) are retained at audit precision; these are cached author predictions, not new model runs.}\label{tab:replay}
\begin{tabular}{llrrr}
\toprule
Data & Prediction & AUC & AP & PR trap. \\
\midrule
UCF & Raw & 72.7904 & 18.2100 & 19.5168 \\
UCF & Original refinement & 80.2757 & 27.1911 & 27.0773 \\
XD & Raw & 80.6145 & 53.0951 & 57.2054 \\
XD & Original refinement & 85.3642 & 62.0045 & 62.0058 \\
\bottomrule\end{tabular}\end{table}

The stored replay reports include independent agreement with the pinned label function and scikit-learn ranking calculations to $10^{-10}$. Those full-frame reports were inspected for this revision; the newly recomputed sampled-anchor analyses are distinct. The full replay table is explicitly historical, not newly rerun GPU inference. All ten refinement-neighbor identities per anchor were checked against source raw scores. The main sampled reference uses only unrefined scores, so refined full-frame results are not its comparator.

The original numerical path was fixed before quality inspection. A mathematically equivalent stable-softmax path changed refined scores by at most $3.33\times10^{-16}$ but altered some tie ordering and aggregate metrics. Both outputs remain archived. We claim reproduction of the published rounded primary values, not bitwise identity with historical saved scalars. No jitter was introduced to improve agreement.

\section{Reproduction and version record}
The pilot-completion ledger held or estimated \$0.110546310588 across 1,602 attempts, including two earlier synthetic connectivity calls; 24 invalid requests retained unknown reservations. This is a historical accounting snapshot, not an invoice or current account balance. No API calls were made for this revision.

The revised pilot evaluator reproduced the saved 20 cells and 44 comparisons exactly. A separate standard-library rank-sum/threshold implementation checked metrics and group-bootstrap differences independently. Constants were also checked against their analytic formulas. The matched LAVAD reference verified 40 cache-file hashes and all 400 dataset/video/anchor joins. New intervals were recomputed rather than borrowed from another comparison. Synthetic CPU/cache tests are implementation checks and contribute no scientific result.

The retained research code supports two distinct operations: aggregate-only paper generation and metric recomputation from private inputs. These programs and their input files are not included in the submission source. Within the research workspace, the following commands run from its \texttt{typed-vad} directory; \texttt{RUN\_QWEN} and \texttt{RUN\_JEV} must name the complete pilot journal directories. A fresh output directory preserves prior analyses:
\begin{lstlisting}[language=bash]
OUT=$(mktemp -d reports/pilot-recheck-XXXXXX)
python3 scripts/evaluate_pilot.py --data .local/pilot-v1 \
  --run "$RUN_QWEN" --run "$RUN_JEV" \
  --output "$OUT/pilot-recomputed.json"
python3 analysis/v1_revision/analyze.py --output-dir "$OUT"
python3 paper/arxiv-v1-final/build.py
PYTHONPATH=src python3 -m unittest discover -s tests -v
python3 -m unittest discover -s analysis/v1_revision -p 'test*.py' -v
\end{lstlisting}
Private-input reanalysis requires the applicable third-party permissions as well as the original journals. The analysis refuses to overwrite its versioned outputs and reads the original pilot paths fixed in its code. The retained aggregate snapshot can regenerate the paper without private text but cannot recompute metrics. In contrast, the submission source only typesets the already computed numbers and vector figures; it needs no Python program, model, credential or private URL. A local clean-source rebuild does not establish arXiv server compilation or acceptance.

This revision added the two constant references, exact matched historical scores, all-issues Choice and input-length audits, and their own exploratory intervals after seeing the pilot. It corrected the description of validation tolerances and excluded confidence/maximum differences from violation accounting. Original prompts, failed outputs, closed journals and frozen splits were preserved. The separate remaining-group protocol and environment were not modified.
\end{document}